\documentclass[10pt,twocolumn,letterpaper]{article}

\usepackage{cvpr}              

\usepackage{algorithm}
\usepackage{algorithmic}
\usepackage{amsmath}
\usepackage{booktabs}
\usepackage{multirow}
\usepackage{pifont}
\usepackage{graphicx}
\usepackage{subcaption}
\usepackage{indentfirst}

\definecolor{cvprblue}{rgb}{0.21,0.49,0.74}
\usepackage[pagebackref,breaklinks,colorlinks,allcolors=cvprblue]{hyperref}

\newcommand{\name}{\textbf{\textsc{Surgeon}}}
\newcommand{\namewithspace}{\textbf{\textsc{Surgeon }}}

\newcommand\mkk[1]{\textcolor{black}{#1}}

\makeatletter
\renewcommand{\@fnsymbol}[1]{*} 
\makeatother

\def\paperID{12420} 
\def\confName{CVPR}
\def\confYear{2025}

\title{\name: \\Memory-Adaptive Fully Test-Time Adaptation via Dynamic Activation Sparsity}

\author{
    Ke Ma$^{1,2}$ \quad Jiaqi Tang$^{3}$ \quad Bin Guo$^{1}$\thanks{Corresponding authors. \tt guob@nwpu.edu.cn, yunhao@tsing hua.edu.cn} \quad Fan Dang$^{4}$ \quad Sicong Liu$^{1}$ \quad Zhui Zhu$^{2}$ \\
    \quad Lei Wu$^{1}$ \quad Cheng Fang$^{1}$ \quad Ying-Cong Chen$^{3}$ \quad Zhiwen Yu$^{1,5}$ \quad Yunhao Liu$^{2}$\textsuperscript{*} \\
    $^1$Northwestern Polytechnical University \quad $^2$Tsinghua University \\ $^3$The Hong Kong University of Science and Technology \\
    $^4$Beijing Jiaotong University \quad $^5$Harbin Engineering University \\
    \\
    \vspace{-0.1in}
    Project Page: \url{https://github.com/kadmkbl/SURGEON}
    \vspace{-0.1in}
}

\begin{document}
\maketitle
\begin{abstract}


Despite the growing integration of deep models into mobile terminals, the accuracy of these models declines significantly due to various deployment interferences.
Test-time adaptation (TTA) has emerged to improve the performance of deep models by adapting them to unlabeled target data online.
Yet, the significant memory cost, particularly in resource-constrained terminals, impedes the effective deployment of most backward-propagation-based TTA methods.
To tackle memory constraints, we introduce \name, a method that substantially reduces memory cost while preserving comparable accuracy improvements during fully test-time adaptation (FTTA) without relying on specific network architectures or modifications to the original training procedure.
Specifically, we propose a novel dynamic activation sparsity strategy that directly prunes activations at layer-specific dynamic ratios during adaptation, allowing for flexible control of learning ability and memory cost in a data-sensitive manner.
Among this, two metrics, Gradient Importance and Layer Activation Memory, are considered to determine the layer-wise pruning ratios, reflecting accuracy contribution and memory efficiency, respectively.
Experimentally, our method surpasses the baselines by not only reducing memory usage but also achieving superior accuracy, delivering SOTA performance across diverse datasets, architectures, and tasks.

\end{abstract}    

\section{Introduction}
\label{sec_introduction}

In recent years, with the progression of artificial intelligence and Internet of Things (IoT) technologies, the integration of deep models into mobile terminals has become increasingly prevalent~\cite{liu2021adaspring,yang2022edgeduet}.
However, due to various types of interference and distribution shifts~\cite{pan2009survey} in deployment environments, such as changing weather conditions~\cite{sakaridis2021acdc} and varying sensor parameters~\cite{kocabas2021spec}, the accuracy of deep models can significantly decline during inference.

To address the aforementioned issue, test-time adaptation (TTA)~\cite{wang2021tent,wang2022continual,liang2024comprehensive} has emerged as an effective strategy to improve the performance of deep models by adapting them to unlabeled target data online. It proves to be particularly advantageous in situations where there is a significant disparity between the test and training data, or when real-time adaptation to fluctuating test environments is necessary.
However, in some practical scenarios, the original training procedure cannot be modified due to concerns over privacy and the storage of training data. To address this, \textit{fully test-time adaptation} (FTTA) has been proposed~\cite{niu2023towards, zhao2023delta} to allow deep models to adapt online without altering the original training procedure.

\begin{figure}[t]
    \centering
    \includegraphics[width=0.48\textwidth]{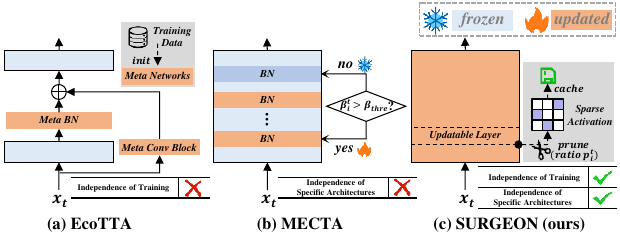}
    \caption{\textbf{The problem of memory-efficient TTA methods}. \textbf{(a)} \textbf{EcoTTA} introduces meta networks to adapt the frozen backbone but requires modifications to the original training procedure to warm up these additional blocks. \textbf{(b)} \textbf{MECTA} establishes the updating criterion based on BN layers. \textbf{(c)} Our method, \namewithspace prunes activations at layer-specific dynamic ratios without relying on specific architectures or modifications to the training procedure.
    \vspace{-0.3in}
    }
    \label{fig_prob_comparison}
\end{figure}

Nevertheless, memory cost becomes a major obstacle to deploying most backward-propagation-based TTA methods~\cite{liu2021ttt++,chen2022contrastive,niu2022efficient,yuan2023robust,dobler2023robust}, particularly in memory-constrained terminals.
In these methods, memory usage primarily arises from the storage of activations necessary for gradient calculation and weight updates during backpropagation (see \S \ref{subsec_optimization}).
To address memory limitations in TTA, EcoTTA~\cite{song2023ecotta} is proposed as a solution where plug-and-play meta networks are designed additionally. Only these meta networks are updated while the backbone of the network is frozen during adaptation, effectively avoiding the memory usage of storing activations for the frozen layers.
However, this approach necessitates modifications to the original training procedure for initializing the additional blocks, making it impractical to deploy in FTTA scenarios, as shown in Figure \ref{fig_prob_comparison} (a).
MECTA~\cite{hong2023mecta} has been introduced as a method suitable for FTTA, which effectively reduces memory cost at the batch, channel, and layer levels. However, as shown in Figure \ref{fig_prob_comparison} (b), this method relies on Batch Normalization (BN) layers to establish the updating criterion, limiting its applicability to transformer-based architectures~\cite{liu2021swin}.
Therefore, previous methods have significant limitations regarding the training procedure or network architectures.

To deal with the above problem, we first declare the optimization goal, which is to achieve the optimal trade-off between minimizing memory cost and maximizing accuracy during adaptation. Upon deeper analysis, we find that the optimization bottleneck is the memory usage of activations that are cached for gradient calculation and weight updates during backpropagation ($\S$ \ref{subsec_optimization}).

\mkk{To this end, we propose \name, an effective and easily deployable strategy for memory-efficient FTTA.
It implements layer-wise pruning of activations across different layers during adaptation, without dependency on specific architectures or modifications to the original training procedure, as shown in Figure \ref{fig_prob_comparison} (c).
This strategy precisely controls the learning capacity and memory cost of different layers.
Specifically, to determine the layer-wise activation pruning ratios, our method utilizes two key metrics: Gradient Importance ($G$) and Layer Activation Memory ($M$), which respectively characterize the accuracy contribution on the current data and the efficiency of activation memory usage.
In contrast to existing techniques, such as layer freezing~\cite{yuan2022layer,li2023smartfrz} and methods employing a global static pruning ratio for activations~\cite{jiang2022back,chen2023dropit}, our proposed \textit{dynamic activation sparsity} offers greater adaptability in a data-sensitive manner.}
Experimentally, \namewithspace achieves SOTA performance in both accuracy and memory cost compared to the baselines across diverse datasets, network architectures (convolution-based and transformer-based models), and tasks (image classification and semantic segmentation).
Our contributions are as follows:
\begin{itemize}
    \item We propose \name\ for memory-adaptive fully test-time adaptation, which optimizes memory cost at each layer while maintaining comparable accuracy, without being constrained by specific architectures or modifications to the original training procedure.
    \item We propose a novel memory-efficient strategy, dynamic activation sparsity, which utilizes layer importance metrics that consider accuracy contribution and memory efficiency. This allows for more granular and flexible control over the learning ability and memory cost of different layers in dynamic FTTA scenarios.
    \item Our method achieves SOTA performance in both accuracy and memory cost across a combination of different datasets, network architectures, and tasks.
\end{itemize}
\section{Related Work}
\label{sec_relatedwork}

\paragraph{Memory-Efficient Fully Test-Time Adaptation}
To mitigate accuracy degradation caused by distribution shift, test-time adaptation (TTA)~\cite{sun2020test,liu2021ttt++,chen2022contrastive,liu2023deja,yuan2023robust} adapts deep models to unlabelled data online.
Recently, fully test-time adaptation (FTTA) has been developed, allowing the adaptation of pre-trained models~\cite{croce2021robustbench} without modifying the original training procedure~\cite{wang2021tent, niu2023towards, zhao2023delta}.

However, in resource-constrained terminals, the memory cost of backward propagation remains a major challenge for deployment~\cite{liu2023enabling, niu2024test}.
To address memory limitations in TTA, Song et al. propose EcoTTA~\cite{song2023ecotta}, which enhances memory efficiency by updating plug-in meta networks while keeping the backbone frozen. However, it requires retraining the meta networks for several epochs using the training data before deployment, making it impractical for FTTA scenarios.
Hong et al. introduce MECTA~\cite{hong2023mecta} to reduce memory usage by optimizing batch size, channels, and layers of adaptation without modifying the original training procedure.
Specifically, it utilizes the statistics of BN layers to determine which subset of layers to update.
However, MECTA's reliance on specific layers limits its applicability to transformer-based architectures~\cite{liu2021swin}, which often lack BN layers.
Our method significantly reduces the memory cost of FTTA while maintaining comparable accuracy improvements.
Crucially, it does not rely on specific architectures or modifications to the training procedure, ensuring broader applicability.

\vspace{-0.1in}
\paragraph{Memory-Efficient Strategies for Backpropagation}
Existing works~\cite{cai2020tinytl,wang2022melon,jiang2022back} have identified that significant adaptation memory cost mainly arises from storing activations for gradient calculation and weight updates during backpropagation. Therefore, the key to memory efficiency is reducing the memory usage of activations (see $\S$ \ref{subsec_optimization}).

One approach is to incorporate additional plug-and-play blocks~\cite{yang2022rep,he2022towards,lian2022scaling} into the network, updating only these blocks during adaptation while keeping the backbone frozen.
The limitation of this approach in FTTA is the necessity of warming up the additional blocks using the training data.
Besides, gradient checkpointing (GC)~\cite{chen2016training} is a straightforward approach  that avoids storing activations for a subset of layers and instead recomputes them before gradient calculation.
However, since it does not alter the gradient values, its accuracy in TTA scenarios is limited.
Another similar approach is layer freezing, where only a subset of network layers is adapted while the rest completely skip gradient calculation and weight updates. To select updated layers, some methods still require training procedure modifications for contribution analysis~\cite{lin2022device} or additional policy networks~\cite{li2023smartfrz}, while others use fixed epoch intervals to progressively freeze layers~\cite{yuan2022layer} or freeze converged layers by analyzing outputs across epochs~\cite{bragagnolo2022update}.

Recently, some works~\cite{jiang2022back,chen2023dropit} explore direct pruning of activations across all layers during adaptation. Compared to layer freezing, this activation sparsity strategy allows for finer adjustment of the memory cost via customized pruning ratios. However, existing methods use a global static sparsity, discarding the unique and varying accuracy contributions of different layers~\cite{lee2023surgical}, thus leaving room for accuracy improvement in dynamic FTTA scenarios.

\section{Preliminary}
\label{sec_preliminary}

\subsection{Problem Definition}
\label{subsec_problem}

In fully test-time adaptation (FTTA), we refer to the problem definition from previous works~\cite{wang2021tent}. Let $M_W$ denote the deep models trained on source training data $D_{train}=\{(x,\dot{y}) \sim p_{s}\}$ with weights $W$, where ${x}$ and ${\dot{y}}$ represent the input and ground truth, respectively.
The goal of FTTA is to adapt $M_W$ to target test data $D_{test}=\{(x) \sim p_{t}\}$ online in an unsupervised manner, mitigating performance degradation from distribution shifts ($p_{s} \neq p_{t}$). The target data distribution $p_{t}$ can change over time.
Additionally, compared to general TTA, modifying the original training procedure is not allowed in FTTA~\cite{niu2023towards}.

Generally, the adaptation procedure is implemented based on backward propagation after forward propagation on a test batch~\cite{chen2022contrastive,niu2022efficient,yuan2023robust}, as shown below,
\vspace{-0.06in}
\begin{equation}
\label{equ_forward}
y=f_{W_l}\left(\ldots f_{W_2}\left(f_{W_1}(x)\right) \ldots\right), L=\mathcal{L}(y),
\end{equation}
\begin{equation}
\label{equ_backward}
\begin{split}
\Delta A_{i}=\frac{\partial L}{\partial A_{i}}=\frac{\partial L}{\partial A_{i+1}} \frac{\partial A_{i+1}}{\partial A_i}=\frac{\partial L}{\partial A_{i+1}} W_i^T, \\
\Delta W_{i}=\frac{\partial L}{\partial W_{i}}=\frac{\partial L}{\partial A_{i+1}} \frac{\partial A_{i+1}}{\partial W_i}=A_i^T \frac{\partial L}{\partial A_{i+1}},
\end{split}
\end{equation}
where Eq.\eqref{equ_forward} and Eq.\eqref{equ_backward} illustrate the forward propagation of a network with $l$ layers and the backward propagation of a single linear layer without bias.
The formulations for different layer types are provided in the Appendix for further details.
Here, $\mathcal{L}$ denotes the loss (e.g., entropy minimization~\cite{grandvalet2004semi}), ${f_{W_{i}}}$ is the function of the $i$-th layer with weights $W_{i}$, and $A_{i+1}=f_{W_{i}}(A_{i})=W_{{i}}A_{i}$ represents the activations (i.e., output) of the $i$-th layer.

\subsection{Optimization Objective}
\label{subsec_optimization}


Our optimization objective is to achieve the optimal balance between minimizing memory cost and maximizing accuracy improvement during FTTA, as shown in Eq.~\eqref{equ_goal}.
\vspace{-0.06in}
\begin{equation}
\label{equ_goal}
\text{min} \quad \alpha \cdot \mathbf{Memory} - \beta \cdot \mathbf{Accuracy},
\end{equation}
where $\alpha$ and $\beta$ regulate the trade-off between memory usage and accuracy.
When updating the $i$-th layer, activations $A_{i}$, weights $W_{i}$, gradients $\Delta A_{i}$ and $\Delta W_{i}$ are cached for backpropagation (see Eq.~\eqref{equ_backward}).
It has been found that the primary memory bottleneck is the activations, $A_{i}$, which can consume over 80\% or even more of adaptation memory usage, as analyzed in the Appendix.
Therefore, \textit{the key to memory efficiency is reducing the activation memory cost}.
\vspace{-0.05in}
\section{Methodology}
\label{sec_methodology}

\begin{figure*}[t]
\centering
  \includegraphics[width=0.9\textwidth]{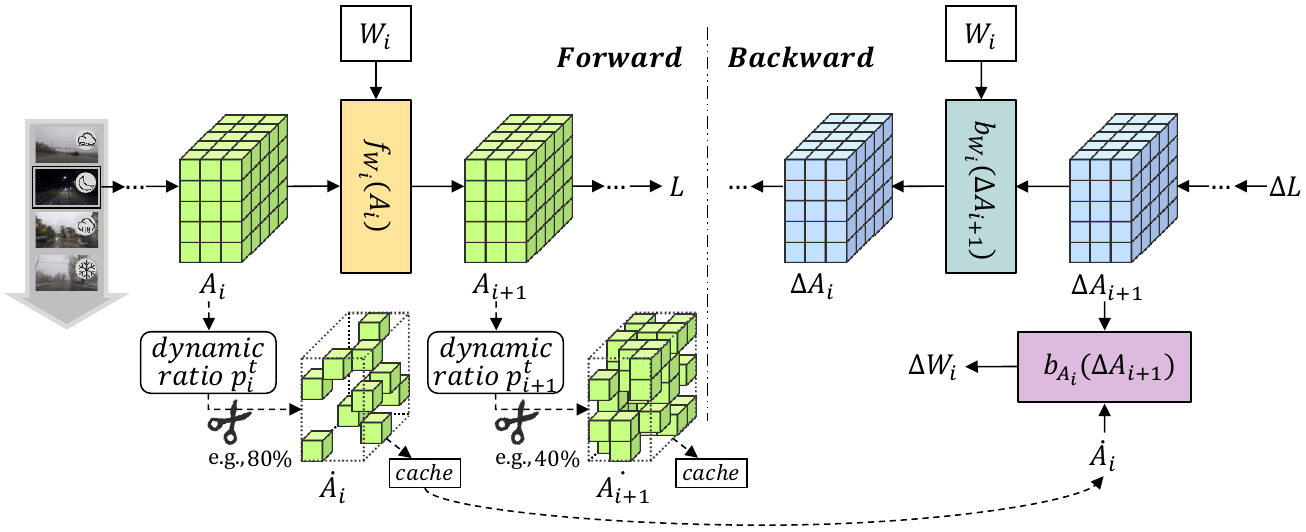}
  \caption{\namewithspace prunes activations at layer-specifc dynamic ratios in a data-sensitive manner during adaptation. In forward propagation, it prunes activations ($A_{i} \rightarrow \dot{A_i}$) before caching them into memory. In backward propagation, these sparse activations are used to calculate the weight gradients $\Delta W_i$ (see Eq.~\eqref{equ_backward}).
  By employing dynamic activation sparsity, \namewithspace substantially reduces the memory cost of adaptation while maintaining comparable accuracy in dynamic FTTA scenarios.
  }
\label{fig_framework}
\end{figure*}

\paragraph{Motivation}
To achieve the objective (Eq.~\eqref{equ_goal}), we propose a novel memory-efficient strategy, dynamic activation sparsity.
By pruning cached activations $A_{i}$ at \textbf{layer-specific} and \textbf{dynamic} ratios in a data-sensitive manner, this strategy precisely regulates both activation memory consumption and weight gradients $\Delta W_{i}$ (see Eq.\eqref{equ_backward}) during adaptation.
This is motivated by the finding that different layers exhibit distinct activation usage and contribute variably to accuracy under changing data distributions~\cite{lee2023surgical}.
Compared to using a global static activation pruning ratio~\cite{jiang2022back,chen2023dropit}, dynamic activation sparsity accounts for inter-layer differences and offers greater adaptability to dynamic FTTA scenarios.

\vspace{-0.15in}
\paragraph{Overview}
Initially, to drive dynamic activation pruning across layers, it is essential to evaluate and prioritize the importance of each layer. In $\S$ \ref{subsec_MGI}, we introduce two metrics: Gradient Importance ($G$) and Layer Activation Memory ($M$) to evaluate layer importance in terms of accuracy contribution and memory efficiency, respectively.
Based on this importance evaluation, layer-wise dynamic activation pruning ratios are determined, and the workflow for FTTA with dynamic activation sparsity is proposed in $\S$ \ref{subsec_workflow}.
Ultimately, \namewithspace effectively balances memory cost and accuracy in FTTA without relying on specific architectures or modifications to the original training procedure.

\subsection{Evaluating Importance for Activation Pruning}
\label{subsec_MGI}

\paragraph{Gradient Importance}
Inspired by ideas that different layers contribute unequally to adaptation under varying data distributions~\cite{lee2023surgical}, implementing layer-specific dynamic activation pruning based on accuracy contribution can further enhance accuracy compared to global static pruning.
To achieve this, we first need to quantify the accuracy contribution of different layers.
Here, the weight gradients are utilized as the reference metric.
The insight is that layers with larger gradients have a more significant influence on model predictions and loss reduction, indicating their potential to capture richer knowledge and achieve greater accuracy gains on the current data~\cite{evci2020rigging}.
The gradient importance of the $i$-th layer is expressed by Eq.~\eqref{equ_gradientimportance},
\begin{equation} \label{equ_gradientimportance} 
\Delta w_i=\frac{\partial L}{\partial w_i}, \quad G_i=\sqrt{\frac{\sum_{j=1}^{N_i}\left(\Delta w_j\right)^2}{N_i}}, \end{equation}
where $\Delta w_{j}$ refers to the weight gradients of the neuron $j$ in the $i$-th layer and $N_{i}$ denotes the number of neurons.
In Eq.~\eqref{equ_gradientimportance}, $G_{i}$, the average gradient of individual neurons within the $i$-th layer serves as the quantification metric for layer gradient importance, as it reflects the layer's impact on loss reduction and accuracy contribution during adaptation.

\vspace{-0.06in}
\paragraph{Layer Activation Memory Importance}
Besides gradient importance, we further consider activation memory efficiency for layer importance evaluation, which is quantified by the size of activation $A_i$.
This metric directly reflects the activation memory usage needed for updating layer weights $W_i$.
Incorporating this metric for layer-wise pruning can further improve memory efficiency during adaptation.
Eq.~\eqref{equ_memoryefficiency} shows the activation memory importance of the $i$-th layer,
\begin{equation} \label{equ_memoryefficiency} m_i=\mathbf{size}(A_i), \quad M_i=-\log\left(\frac{m_i}{\sum_{i=1}^lm_i}\right), \end{equation}
where $m_i$ denotes the size of the activations $A_i$, and $M_i$ is calculated using the logarithm of the ratio between $m_i$ and the total activation size across all layers.

\paragraph{Combination of Importance Metrics}
Finally, we combine these two layer importance metrics as shown in Eq.~\eqref{equ_MGI},
\begin{equation}
\label{equ_MGI}
I_i=  \mathbf{Norm}(M_i) \times \mathbf{Norm}(G_i),
\end{equation}
where $M_i$ and $G_i$ are scaled to [0,1] using Max Normalization for uniformity. Finally, the layer importance indicator, $I_i$, integrates the accuracy contribution and memory efficiency during adaptation to inform decisions on determining layer-wise activation pruning ratios.

\subsection{Workflow of Dynamic Activation Sparsity}
\label{subsec_workflow}

Once we have evaluated the layer importance using the combined metric $I_{i}$ via Eq.~\eqref{equ_MGI}, the next step is to determine the activation pruning ratios across different layers.
Eq.~\eqref{equ_pruningratio} illustrates the process of converting $I_i$ into layer-wise pruning ratios at the current data batch $t$,
\begin{equation}
\label{equ_pruningratio}
p^t_i=1-\frac{I^t_i}{\max _{i \in\{1,2, \ldots, l\}}\left(I^t_i\right)},
\end{equation}
where the activation pruning ratio $p^t_i$ of the $i$-th layer is calculated as 1 minus the ratio of $I^t_i$ to the maximum value of the importance indicators across all layers.
Overall, \namewithspace encourages assigning \textit{low pruning ratios to activations $A_{i}$ when layer weights $W_{i}$ have large gradients} due to these layers' significant accuracy contribution on current data. Conversely, it tends to assign \textit{high pruning ratios to activations $A_{i}$ with large sizes} to further enhance memory efficiency during adaptation.

Figure \ref{fig_framework} illustrates the layer-specific dynamic activation pruning process during adaptation. In the forward propagation, the activation $A_{i}$ is pruned and cached immediately after $A_{i+1}$ is computed. In the backward propagation, the cached sparse activation $\dot{A_{i}}$ is then used to calculate the weight gradients $\Delta W_i$ (see Eq.~\eqref{equ_backward}).

\paragraph{Algorithm} Finally, we summarize the workflow of our methods in Algorithm \ref{algorithm_workflow}.
Following Jiang et al.~\cite{jiang2022back}, to genuinely reduce memory usage within the processors, we cache the pruned activations $\dot{A_i}$ as two components (line 7): $\hat{A_i}$ and $idx_i$. $\hat{A_i}$ is a one-dimensional vector that stores only the non-zero elements of $\dot{A_i}$, while $idx_i$ records their indices using 1 bit per element. In the backward propagation, the pruned activations $\dot{A_i}$ are first reshaped (line 11) and then used for gradient calculation (line 12).

\vspace{-0.15in}
\paragraph{Memory Efficiency of Layer Importance Calculation}
Notably, before the adaptation process, an additional forward-backward process is required initially to calculate the importance metrics $G_{i}$ and $M_{i}$ (line 2).
\mkk{To reduce the memory cost of the additional process and ensure that it does not exceed the peak memory usage of the following adaptation process, we can use two effective strategies: (i) randomly sampling a small subset of the current data batch, and (ii) applying a high global static pruning ratio (e.g., 90\%) to acitvations. Details can be seen in the Appendix.}

\begin{algorithm}[tb]
\caption{FTTA with Dynamic Activation Sparsity}
\label{algorithm_workflow}
\textbf{Require}: Network with $l$ layers and weights $\{W_i\}$, test data with total $n$ batches.
\begin{algorithmic}[1]
\FOR{each batch $t \in \{1, 2, \ldots, n\}$}
    \STATE Calculate pruning ratios $\{p^t\}$ across all layers via an additional forward-backward process \\
    // Forward Propagation
    \FOR{$i \in \{1, 2, \ldots, l\}$}
        \STATE $A_i \gets \text{Get the input of the $i$-th layer}$
        \STATE $A_{i+1} \gets \text{Calculate the output of the $i$-th layer}$
        \STATE $\dot{A_i} \gets \text{Prune the activations $A_i$ with ratio $p^t_i$}$
        \STATE $\hat{A_i}, idx_i \gets$ Decompose $\dot{A_i}$ into the non-zero elements and the index, and cache them into memory
    \ENDFOR
    \STATE Calculate Loss \\
    // Backward Propagation
    \FOR{$i \in \{l, l-1, \ldots, 1\}$}
        \STATE $\dot{A_i} \gets Reshape(\hat{A_i}, idx_i)$
        \STATE $\Delta A_i,\Delta W_i \gets \text{Calculate the activation gradients}$ \\ \text{and weight gradients via Eq.~\eqref{equ_backward}} using $\dot{A_i}$
    \ENDFOR
    \STATE Update the weights with weight gradients $\{\Delta W_i\}$
\ENDFOR
\end{algorithmic}
\end{algorithm}
\section{Experiments}
\label{sec_exp}

\subsection{Experimental Settings}
\label{subsec_expdetails}
\paragraph{Datasets and Architectures}
We employ two downstream tasks: image classification and semantic segmentation.
\textbf{For image classification}, we use CIFAR10-C, CIFAR100-C and ImageNet-C~\cite{hendrycks2019benchmarking} as the out-of-distribution test data originating from 15 types of corruptions at the severity level 5.
For the network architectures, we utilize WideResNet-28~\cite{zagoruyko2016wide} trained on CIFAR10, ResNeXt-29~\cite{xie2017aggregated} trained on CIFAR100, and ResNet-50 (AugMix)~\cite{he2016deep} trained on ImageNet, with all these pre-trained weights from RobustBench~\cite{croce2021robustbench}.
\textbf{For semantic segmentation}, we use ACDC~\cite{sakaridis2021acdc} as the out-of-distribution test data, which contains images collected in four different conditions (i.e., rain, snow, fog, night). For the network architectures, we utilize ResNet-50 (from the backbone of DeeplabV3+~\cite{chen2018encoder}) and transformer-based Segformer-B5~\cite{xie2021segformer} both trained on Cityscapes~\cite{cordts2016cityscapes}, with pre-trained weights from RobustNet repository~\cite{choi2021robustnet} and CoTTA repository~\cite{wang2022continual}, respectively.

Following CoTTA, we use the same test sequence for both image classification and semantic segmentation experiments.
The test batch size is set to 200 for CIFAR, 64 for ImageNet, and for semantic segmentation, 2 for ResNet and 1 for Segformer.
Further details on hyperparameters can be found in the Appendix.

\begin{table*}[t]
  \caption{Online error (\%) and cache size (MB) for TTA on CIFAR-to-CIFAR-C with a batch size of 200. Results are obtained on WideResNet-28 for CIFAR10-C and ResNeXt-29 for CIFAR100-C. The \textbf{best} and \underline{second best} scores are highlighted. ``Original'' refers to whether the method requires altering the original training procedure.}
  \centering
  \Huge
  \resizebox{1.0\linewidth}{!}{
    \begin{tabular}{l|l|c|ccccccccccccccc|c|c}
    \toprule
    \multicolumn{1}{l|}{\multirow{2}[2]{*}{Datasets}} & \multirow{2}[2]{*}{Methods} & \multirow{2}[2]{*}{Original} & \multicolumn{15}{c|}{Time Stamp $\xrightarrow{\hspace*{25cm}}$}                                                                          & \multirow{2}[2]{*}{Mean (\%) $\downarrow$} & {\multirow{2}[2]{*}{Cache (MB) $\downarrow$}} \\
          & \multicolumn{1}{l|}{} & \multicolumn{1}{c|}{} &  {Gauss.} &  {Shot} &  {Impul.} &  {Defoc.} &  {Glass} &  {Moti.} &  {Zoom} &  {Snow} &  {Frost} &  {Fog} &  {Brigh.} &  {Contr.} &  {Elast.} &  {Pixel.} & {Jpeg} &       &  {}\\
    \midrule
    \multicolumn{1}{l|}{\multirow{14}[2]{*}{CIFAR10-C}} & Source~\cite{zagoruyko2016wide} & \ding{55} & 72.3  & 65.7  & 72.9  & 46.9  & 54.3  & 34.8  & 42.0    & 25.1  & 41.3  & 26.0  & 9.3   & 46.7  & 26.6  & 58.5  & 30.3  & 43.5  & {125}\\         
          & BN-stat~\cite{schneider2020improving} & \ding{55} & 28.3  & 26.0  & 36.2  & 12.6  & 34.9  & 13.9  & 12.0  & 17.5  & 17.6  & 14.9  & 8.2   & 13.0  & 23.5  & 19.5  & 27.2  & 20.4  & {125}\\
          \cmidrule{2-20}
          & Full Tuning~\cite{lee2023surgical} & \ding{55} & 26.8  & 22.5  & 30.9  & 12.0  & 31.4  & 14.0  & 12.0  & 17.3  & 16.5  & 15.6  & 9.8   & 13.3  & 21.5  & 16.9  & 22.5  & 18.9  & {3697}\\
          & TENT~\cite{wang2021tent}  & \ding{55} & 26.1  & 21.3  & 29.6  & 11.8  & 30.6  & 13.9  & 11.6  & 17.1  & 16.4  & 15.9  & 9.5   & 13.6  & 22.2  & 17.3  & 21.9  & 18.6  & {1762}\\
          & CoTTA~\cite{wang2022continual} & \ding{55} & 24.3  & 21.3  & 26.6  & 11.6  & 27.6  & 12.2  & 10.3  & 14.8  & 14.1  & 12.4  & 7.5   & 10.6  & 18.3  & 13.4  & 17.3  & \textbf{16.2}  & {3697}\\
          \cmidrule{2-20} 
          & EcoTTA~\cite{song2023ecotta} & \ding{51} & 23.5  & 18.5  & 26.1  & 11.4  & 29.3  & 14.1  & 11.5  & 15.7  & 14.4  & 13.6  & 8.6  & 12.1  & 19.4  & 15.2  & 19.6  & \underline{16.8}  & 1200   \\
          \cmidrule{2-20} 
          & MECTA~\cite{hong2023mecta} & \ding{55} & 28.5  & 20.6  & 28.8  & 15.4 & 32.3  & 15.7  & 12.4  & 19.8  & 16.4  & 15.9 & 8.7   & 14.5  & 22.0  & 19.2  & 22.5  & 19.5 & {348}\\
          & +EATA~\cite{niu2022efficient} & \ding{51} & 27.1  & 19.0 & 27.3  & 15.6  & 31.2  & 15.7  & 12.3  & 18.5  & 15.9  & 15.1  & 9.2   & 14.0  & 20.6  & 18.0  & 20.6  & 18.6  & {373}\\
          \cmidrule{2-20} 
          & \name & \ding{55} & 27.0  & 22.8  & 31.3  & 11.8  & 31.0  & 13.2  & 10.9  & 16.0  & 15.0  & 13.8  & 8.1  & 11.7  & 20.6  & 16.2  & 22.8  & 18.1  & {325}\\
          & +CSS  & \ding{55} & 26.9  & 22.6  & 31.1  & 11.8  & 30.8  & 13.0  & 10.9  & 16.1  & 15.0  & 13.9  & 8.2  & 11.5  & 20.3  & 16.0  & 21.8  & 18.0  & {312}\\
          & +CSS \& CR & \ding{55} & 25.7  & 21.1  & 28.6  & 11.9  & 28.4  & 13.0  & 10.8  & 15.3  & 14.2  & 13.6  & 8.2   & 11.7  & 18.2  & 14.3  & 18.7  & 16.9  & {581}\\
          \cmidrule{2-20} 
          & \namewithspace (BN) & \ding{55} & 26.5  & 21.6  & 29.9  & 12.0  & 30.3  & 13.5  & 10.9  & 15.4  & 14.5  & 13.9  & 8.4  & 11.7  & 21.0  & 15.2  & 22.2  & 17.8  & {\underline{242}}\\ 
          & + CSS & \ding{55} & 26.0  & 21.1  & 29.4  & 11.9  & 30.2  & 13.4  & 10.9  & 15.6  & 14.9  & 14.5  & 8.3  & 12.4  & 20.8  & 15.3  & 21.0  & 17.7  & {\textbf{237}}\\ 
          & + CSS \& CR & \ding{55} & 25.3  & 20.0  & 27.4  & 12.0  & 28.0  & 13.7  & 11.11 & 15.4  & 14.3  & 14.1  & 8.2   & 11.8  & 19.2  & 13.5  & 18.7  & \underline{16.8}  & {434}\\
    \midrule
    \multicolumn{1}{l|}{\multirow{14}[2]{*}{CIFAR100-C}} & Source~\cite{xie2017aggregated} & \ding{55} & 73.0  & 68.0  & 39.4  & 29.4  & 54.1  & 30.8  & 28.8  & 39.5  & 45.8  & 50.3  & 29.5  & 55.1  & 37.2  & 74.7  & 41.2  & 46.5  & {200}\\
          & BN-stat~\cite{schneider2020improving} & \ding{55} & 42.3  & 40.9  & 43.3  & 27.7  & 41.9  & 29.8  & 27.9  & 35.1  & 35.0  & 41.7  & 26.3  & 30.3  & 35.6  & 33.4  & 41.3  & 35.5  & {200}\\
          \cmidrule{2-20}
          & Full Tuning~\cite{lee2023surgical} & \ding{55} & 40.7  & 36.0  & 37.7  & 26.8  & 38.2  & 29.5  & 27.6  & 34.3  & 33.3  & 40.2  & 28.0  & 33.0  & 35.3  & 32.7  & 40.5  & 34.3  & {5403}\\
          & TENT~\cite{wang2021tent}  & \ding{55} & 41.0  & 37.0  & 38.4  & 25.9  & 37.8  & 28.1  & 25.7  & 32.4  & 31.8  & 37.4  & 25.2  & 29.2  & 32.8  & 29.9  & 38.9  & 32.8  & {2725}\\
          & CoTTA~\cite{wang2022continual} & \ding{55} & 40.1  & 37.7  & 39.7  & 26.9  & 38.0  & 27.9  & 26.4  & 32.8  & 31.8  & 40.3  & 24.7  & 26.9  & 32.5  & 28.3  & 33.5  & 32.5  & {5403}\\
          \cmidrule{2-20} 
          & EcoTTA~\cite{song2023ecotta} & \ding{51} & 40.7  & 38.1  & 41.2  & 26.4  & 41.1  & 28.6  & 26.5  & 32.8  & 31.9  & 38.9  & 24.8  & 28.6  & 33.5  & 29.7  & 37.2  & 33.3  & 1050   \\
          \cmidrule{2-20} 
          & MECTA~\cite{hong2023mecta} & \ding{55} & 42.9  & 39.4  & 41.3  & 29.9  & 42.2  & 29.2  & 26.8  & 36.2  & 33.9  & 42.8  & 25.0  & 32.8  & 34.7  & 32.7  & 38.1  & 35.2  & 632   \\
          & +EATA~\cite{niu2022efficient} & \ding{51} & 43.4  & 39.6  & 41.2  & 30.4  & 43.1  & 29.5  & 27.9  & 37.5  & 34.4  & 41.6  & 25.6  & 32.8  & 35.2  & 33.1  & 38.6  & 35.5  & 637    \\
          \cmidrule{2-20} 
          & \name & \ding{55} & 41.7  & 38.4  & 39.9  & 26.3  & 38.5  & 28.3  & 25.9  & 32.4  & 31.8  & 38.9  & 25.1  & 29.1  & 32.7  & 29.9  & 37.8  & 33.1  & 820    \\
          & +CSS  & \ding{55} & 41.1  & 37.2  & 38.6  & 26.2  & 37.2  & 28.0  & 25.5  & 31.5  & 30.3  & 36.8  & 25.0  & 27.9  & 31.2  & 28.0  & 35.5  & 32.1  & 793    \\
          & +CSS \& CR & \ding{55} & 41.1  & 36.9  & 38.4  & 26.2  & 36.6  & 27.9  & 25.4  & 31.2  & 30.0  & 36.9  & 24.9  & 27.9  & 31.2  & 27.7  & 35.0  & \underline{31.8}  & 1600   \\
          \cmidrule{2-20} 
          & \namewithspace (BN) & \ding{55} & 41.6  & 38.1  & 40.1  & 26.3  & 38.6  & 28.0  & 25.6  & 32.5  & 31.6  & 37.4  & 24.5  & 28.0  & 32.3  & 29.3  & 38.2  & 32.8  & \underline{605}  \\ 
          & + CSS & \ding{55} & 40.4  & 35.8  & 38.3  & 26.2  & 37.1  & 27.9  & 25.2  & 31.8  & 30.7  & 35.1  & 25.0  & 27.3  & 31.5  & 28.7  & 37.3  & \underline{31.8}  & \textbf{588}  \\ 
          & + CSS \& CR & \ding{55} & 40.3  & 35.6  & 38.1  & 26.4  & 36.7  & 28.1  & 25.5  & 31.2  & 30.0  & 35.6  & 24.8  & 27.2  & 31.3  & 28.2  & 36.1  & \textbf{31.7}  & 1093  \\
    \bottomrule
    \end{tabular}
    }
  \label{tab_classification}%
\end{table*}%

\begin{table*}[t]
  \caption{Online error (\%) and cache size (MB) for TTA on Cityscapes-to-ACDC. Results are obtained on ACDC using DeeplabV3+ and Segformer-B5. The \textbf{best} and \underline{second best} scores are highlighted.
  }
  \centering
  \Huge
  \resizebox{1.0\linewidth}{!}{
    \begin{tabular}{l|l|cccc|cccc|cccc|cccc|c|c}
    \toprule
    \multirow{3}[4]{*}{Architectures} & \multirow{3}[4]{*}{Methods} & \multicolumn{16}{c|}{Time Stamp $\xrightarrow{\hspace*{25cm}}$}                                                                                   & \multicolumn{1}{c|}{\multirow{3}[4]{*}{Mean (\%) $\downarrow$}} & \multirow{3}[4]{*}{Cache (MB) $\downarrow$} \\
    \multicolumn{1}{l|}{} & \multicolumn{1}{l|}{} & \multicolumn{4}{c|}{1}        & \multicolumn{4}{c|}{4} & \multicolumn{4}{c|}{7} & \multicolumn{4}{c|}{10} &       &  \\
    \multicolumn{1}{l|}{} & \multicolumn{1}{l|}{} &  {Fog} &  {Night} &  {Rain} & {Snow} &  {Fog} &  {Night} &  {Rain} & {Snow} &  {Fog} &  {Night} &  {Rain} & {Snow} &  {Fog} &  {Night} &  {Rain} & {Snow} & &  \\
    \midrule
    \multicolumn{1}{l|}{\multirow{8}[2]{*}{DeeplabV3+}} & Source~\cite{chen2018encoder} & 38.6  & 82.4  & 47.3  & 53.9  & 38.6  & 82.4  & 47.3  & 53.9  & 38.6  & 82.4  & 47.3  & 53.9  & 38.6  & 82.4  & 47.3  & 53.9  & 55.6  &  301    \\
    \multicolumn{1}{l|}{} & BN-stat~\cite{schneider2020improving} & 57.5  & 73.1  & 55.2  & 57.8  & 57.5  & 73.1  & 55.2  & 57.8  & 57.5  & 73.1  & 55.2  & 57.8  & 57.5  & 73.1  & 55.2  & 57.8  & 60.9  & 301    \\
    \cmidrule{2-20}
    \multicolumn{1}{l|}{} & Full Tuning~\cite{lee2023surgical} & 57.0  & 72.5  & 53.3  & 55.5  & 52.8  & 72.1  & 50.8 & 54.4 & 54.7  & 74.5  & 52.8  & 56.9  & 57.6  & 77.0  & 55.7  & 59.6  & 59.4  & 9974   \\
    \multicolumn{1}{l|}{} & TENT~\cite{wang2021tent}  & 57.4  & 73.0  & 54.8  & 57.3  & 55.7  & 72.2  & 53.1  & 55.8  & 54.6  & 71.7  & 51.9  & 54.8  & 53.8  & 71.3  & 51.2  & 54.2  & 58.9  & 4646   \\
    \multicolumn{1}{l|}{} & CoTTA~\cite{wang2022continual} & 38.7  & 84.0  & 47.3  & 53.8  & 38.8  & 84.1  & 47.3  & 53.9  & 38.8  & 84.1  & 47.3  & 53.9  & 38.8  & 84.1  & 47.3  & 53.9  & 56.0  & 9974  \\
    \multicolumn{1}{l|}{} & MECTA~\cite{hong2023mecta} & 56.9  & 75.2  & 52.0  & 55.2  & 54.5  & 74.1  & 50.3  & 53.8  & 53.1  & 74.0  & 50.1  & 53.4  & 53.1  & 74.0  & 50.1  & 53.4  & 58.2  & \underline{1354}   \\
    \cmidrule{2-20}
    \multicolumn{1}{l|}{} & \name  & 51.6  & 72.5  & 51.8  & 54.0  & 48.7  & 71.7  & 48.1  & 50.9  & 47.1  & 71.7  & 46.9  & 49.4  & 46.7  & 72.0  & 46.6  & 49.0  & \underline{54.7}  & 1403   \\
    \cmidrule{2-20}
    \multicolumn{1}{l|}{} & \namewithspace (BN)  & 51.7  & 72.4  & 51.9  & 54.0  & 47.7  & 70.9  & 48.2  & 50.0  & 46.4  & 70.8  & 47.2  & 49.2  & 45.6  & 71.1  & 47.5  & 49.2  & \textbf{54.4}  & \textbf{1300} \\
    \midrule
    \multicolumn{1}{l|}{\multirow{4}[2]{*}{Segformer-B5}} & Source~\cite{xie2021segformer} & 30.9  & 59.7  & 40.3  & 42.2  & 30.9  & 59.7  & 40.3  & 42.2  & 30.9  & 59.7  & 40.3  & 42.2  & 30.9  & 59.7  & 40.3  & 42.2  & 43.3  & 380    \\
    \cmidrule{2-20}
    \multicolumn{1}{l|}{} & CoTTA~\cite{wang2022continual} & 29.1  & 58.8  & 37.6  & 40.3  & 29.1  & 59.0  & 37.3  & 40.3  & 29.1  & 59.0  & 37.2  & 40.3  & 29.2  & 59.0  & 37.2  & 40.3  & \textbf{41.4}  & 2793    \\
    \multicolumn{1}{l|}{} & MECTA~\cite{hong2023mecta} & 30.9  & 59.8  & 40.0  & 42.6  & 33.1  & 63.1  & 41.1  & 45.7  & 35.2  & 66.5  & 46.0  & 48.2  & 37.3  & 69.4  & 47.1  & 51.0  & 47.2  & 380    \\
    \cmidrule{2-20}
    \multicolumn{1}{l|}{} & \name  & 30.9  & 59.7  & 40.0  & 42.2  & 30.4  & 59.1  & 39.1  & 41.7  & 30.0  & 58.7  & 38.3  & 41.3  & 29.7  & 58.3  & 37.9  & 41.0  & \underline{42.2}  & 380    \\
    \bottomrule
    \end{tabular}}
  \label{tab_segmentation}%
  \vspace{-0.1in}
\end{table*}%

\vspace{-0.15in}
\paragraph{Evaluation Metrics}
Following MECTA~\cite{hong2023mecta}, we use three evaluation metrics: (i) \textbf{Online Error (\%)}, which is the average prediction error for each class in each domain. (ii) \textbf{Mean Online Error (\%)}, which is the average of all online errors across domains in the test sequence. (iii) \textbf{Cache Size (MB)}, which indicates the average memory cost of all cached activations during adaptation, ignoring that of weights and gradients because they are linearly related to the number of parameters in the network.

\begin{figure*}[t]
    \centering
    \begin{subfigure}{0.32\textwidth}
        \centering
        \includegraphics[width=\textwidth]{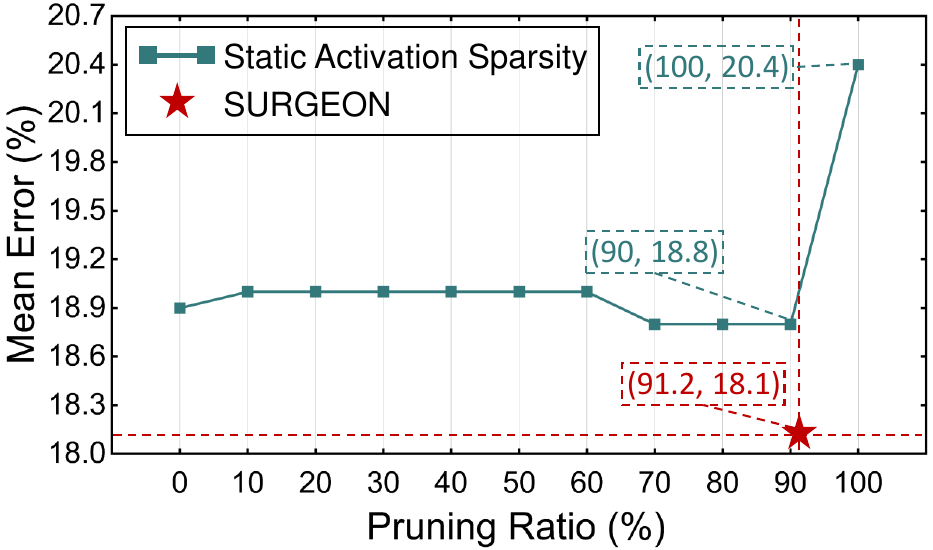}
        \caption{WideResNet-28}
        \label{fig_samecache1}
    \end{subfigure}
    \begin{subfigure}{0.32\textwidth}
        \centering
        \includegraphics[width=\textwidth]{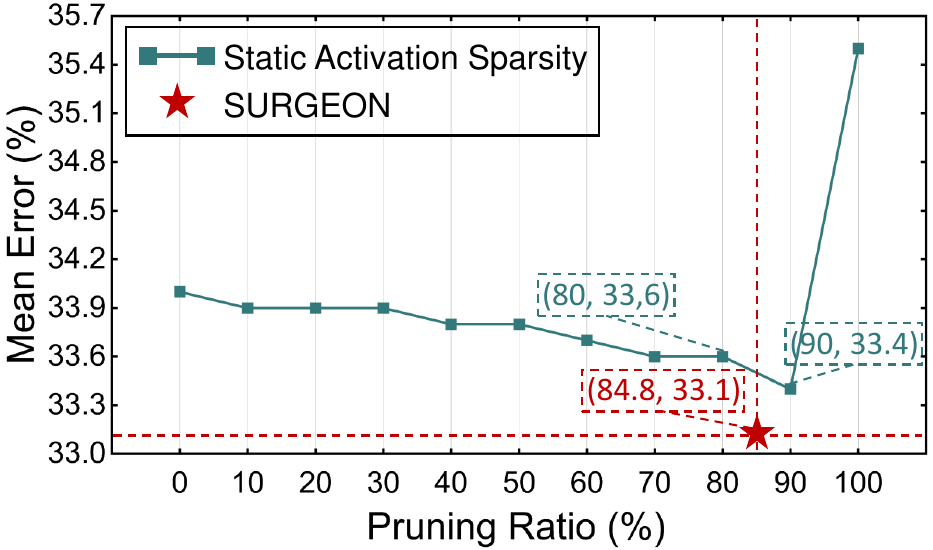}
        \caption{ResNeXt-29}
        \label{fig_samecache2}
    \end{subfigure}
    \begin{subfigure}{0.32\textwidth}
        \centering
        \includegraphics[width=\textwidth]{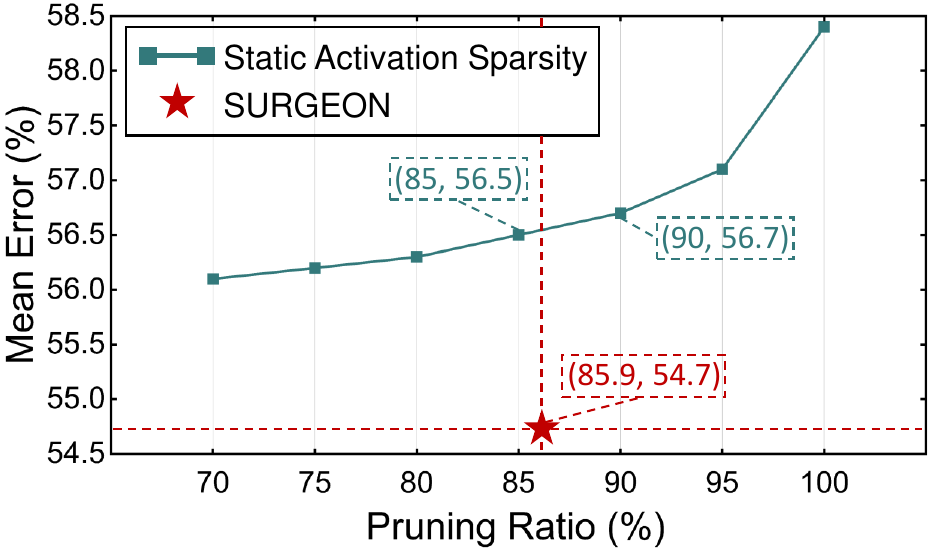}
        \caption{DeeplabV3+}
        \label{fig_samecache3}
    \end{subfigure}
    \caption{Mean online error (\%) under different global static pruning ratios for TTA on three convolutional networks. The pruning ratio of \namewithspace refers to a global static pruning ratio that yields an equivalent cache size.}
    \label{fig_samecache}
\end{figure*}

\begin{table*}[t]
  \caption{Mean online error (\%) and cache size (MB) for TTA with different layer importance metrics. ``w/o $G$ + $M$'' is the TTA method using \textbf{Full Tuning} and ``Ours'' is \name\ with the original design of $I$ (Eq. \eqref{equ_MGI}).}
  \centering \small
    \begin{tabular}{l|cc|cc|cc}
    \toprule
    \multirow{2}[1]{*}{Strategies} & \multicolumn{2}{c|}{WideResNet-28} & \multicolumn{2}{c|}{ResNeXt-29} & \multicolumn{2}{c}{DeeplabV3+} \\
          & Mean (\%) $\downarrow$ & Cache (MB) $\downarrow$ & Mean (\%) $\downarrow$ & Cache (MB) $\downarrow$ & Mean (\%) $\downarrow$ & Cache (MB) $\downarrow$ \\
          \midrule
    w/o $G$ + $M$  & 18.9  & 3697  & 34.3  & 5403  & 59.4  & 9974 \\
    w/o $G$ & 20.3 (1.4 $\uparrow$)  & 1568 (57.6\% $\downarrow$)  & 34.0 (0.3 $\downarrow$)  & 2396 (55.7\% $\downarrow$)  & 56.5 (2.9 $\downarrow$)  & 2385 (76.1\% $\downarrow$)  \\
    w/o $M$ & 17.8 (1.1 $\downarrow$)  & 704 (81.0\% $\downarrow$)   & 32.8 (1.5 $\downarrow$)  & 1108 (79.5\% $\downarrow$)  & 54.8 (4.6 $\downarrow$)  & 1596 (84.0\% $\downarrow$)  \\
    Ours   & 18.1 (0.8 $\downarrow$)  & 325 (91.2\% $\downarrow$)   & 33.1 (1.2 $\downarrow$)  & 820 (84.8\% $\downarrow$)   & 54.7 (4.7 $\downarrow$)  & 1403 (85.9\% $\downarrow$) \\
    \bottomrule
    \end{tabular}
  \label{tab_ablation}%
\end{table*}%

\vspace{-0.15in}
\paragraph{Baselines}
We introduce the following TTA baselines for comparison with \name.
\textbf{Source} indicates the original network without any adaptation.
\textbf{BN-stat}~\cite{schneider2020improving} only updates the running statistics (i.e., mean $\mu$ and deviation $\sigma$) of BN layers during adaptation.
\textbf{TENT}~\cite{wang2021tent} updates both the running statistics and the weights of BN layers via entropy minimization loss~\cite{grandvalet2004semi}.
\textbf{Full-Tuning}~\cite{lee2023surgical} updates the entire network during adaptation.
\textbf{CoTTA}~\cite{wang2022continual} is a SOTA TTA method that utilizes weight-averaged and augmentation-averaged predictions, along with a stochastic restoration mechanism, to enhance its adaptation accuracy.
\textbf{EcoTTA}~\cite{song2023ecotta} is a memory-efficient TTA method that introduces lightweight meta networks to adapt the frozen backbone, but it requires modifications to the original training procedure to warm up these additional blocks.
\textbf{MECTA}~\cite{hong2023mecta} reduces the activation memory cost of TTA in the batch, channel, and layer, but relies on BN layers to build its updating criterion.

\vspace{-0.15in}
\paragraph{Compatibility}
We also aim to prove the compatibility of \name, therefore following MECTA~\cite{hong2023mecta}, we combine two plug-and-play TTA strategies.
(i) \textbf{Certainty-based Sample Selection (CSS)} skips the backpropagation for samples with high entropy values in predictions, as these samples are considered to provide unreliable pseudo-labels, which can hurt adaptation performance~\cite{niu2022efficient}.
(ii) \textbf{Consistency Regularization (CR)} involves adding the discrepancy between model predictions on original test data and augmented test data as regularization to the adaptation loss, aiming to enhance the robustness of pseudo-labels and the performance of unsupervised adaptation~\cite{sohn2020fixmatch}.

\vspace{-0.1in}
\subsection{Evaluation Comparison}
\label{subsec_evaluation}

\paragraph{\namewithspace on CIFAR-C}
As shown in Table \ref{tab_classification}, \textbf{Source} and \textbf{BN-stat} avoid storing activations for backpropagation, so their cache size reflects forward propagation cost, tied to the most memory-intensive layer in the network~\cite{hong2023mecta}.
However, these two baselines also exhibit limited accuracy.
Although \textbf{Full-Tuning} and \textbf{CoTTA} achieve relatively higher accuracy improvements, they also consume the highest memory for storing activations of all layers for backpropagation.
\textbf{TENT} reduces memory usage by only updating BN layers, but it still occupies a significant amount of memory.
\textbf{EcoTTA} efficiently reduces memory cost during adaptation while preserving comparable accuracy via introducing meta networks. However, it necessitates modifications to the original training process, limiting its applicability to FTTA.
\textbf{MECTA} achieves a more significant reduction in activation memory usage, but this comes at the cost of diminished accuracy gains.
For a fair comparison, we create two versions of \name: one that updates all layers of the network and another that only updates BN layers as \namewithspace (BN).
Both versions exhibit outstanding performance in memory reduction and accuracy.

\begin{table}[t]
\caption{Mean online error (\%), cache size (MB) and GFLOPs (per sample) for TTA on ImageNet-to-ImageNet-C. Results are obtained on ResNet-50 (AugMix). ``Original'' refers to whether the method requires altering the original training procedure.}
\resizebox{\linewidth}{!}{
\begin{tabular}{l|c|c|c|c}
\hline
Methods      & Original               & Mean (\%) $\downarrow$  & Cache (MB) $\downarrow$ & GFLOPs $\downarrow$     \\ \hline
Source~\cite{he2016deep}       & \ding{55} & 74.4                & 196          & 4.1          \\
BN-stat~\cite{schneider2020improving}      & \ding{55} & 60.5                & 196          & 4.1          \\ \hline
TENT~\cite{wang2021tent}         & \ding{55} & 55.2(1.2)           & 2714         & \textbf{8.2}          \\
CoTTA~\cite{wang2022continual}        & \ding{55} & 54.4 (3.8)          & 5317         & 103.7        \\
EcoTTA (K=5)~\cite{song2023ecotta} & \ding{51} & {\underline{54.2} (2.5)}           & 1503         & 10.1         \\ \hline
MECTA~\cite{hong2023mecta}        & \ding{55} & 70.1 (10.5)         & \underline{651}    & \textbf{8.2} \\
+ EATA~\cite{niu2022efficient}       & \ding{51} & 64.4 (5.7)          & \textbf{648} & \underline{8.5}    \\ \hline
SURGEON      & \ding{55} & 55.5 (2.6)          & 1125         & 13.4         \\
+ CSS        & \ding{55} & 55.3 (3.9)          & 1084         & 13.4         \\
+ CSS \& CR  & \ding{55} & {\underline{54.2} (2.9)}           & 2201         & 18.0         \\ \hline
SURGEON (BN) & \ding{55} & 55.2 (1.4)          & 912          & 9.7          \\
+ CSS        & \ding{55} & 54.6 (2.0)          & 907          & 9.8          \\
+ CSS \& CR  & \ding{55} & \textbf{53.9} (1.8) & 1834         & 14.2         \\ \hline
\end{tabular}}
\vspace{-0.12in}
\label{tab_imagenet}
\end{table}

\vspace{-0.15in}
\paragraph{\namewithspace on ImageNet-C}
Following CoTTA~\cite{wang2022continual}, we average and report the results over ten diverse corruption sequences for the ImageNet-to-ImageNet-C experiments.
As shown in Table \ref{tab_imagenet}, while \textbf{MECTA} achieves advantages in cache reduction and computational efficiency, it offers limited accuracy improvement.
Both \textbf{EcoTTA} and \namewithspace strike a great balance between accuracy and overhead.
However, our method can be directly applied to FTTA without altering the model's original training procedure.

\begin{figure}[t]
    \centering
    \includegraphics[width=0.48\textwidth]{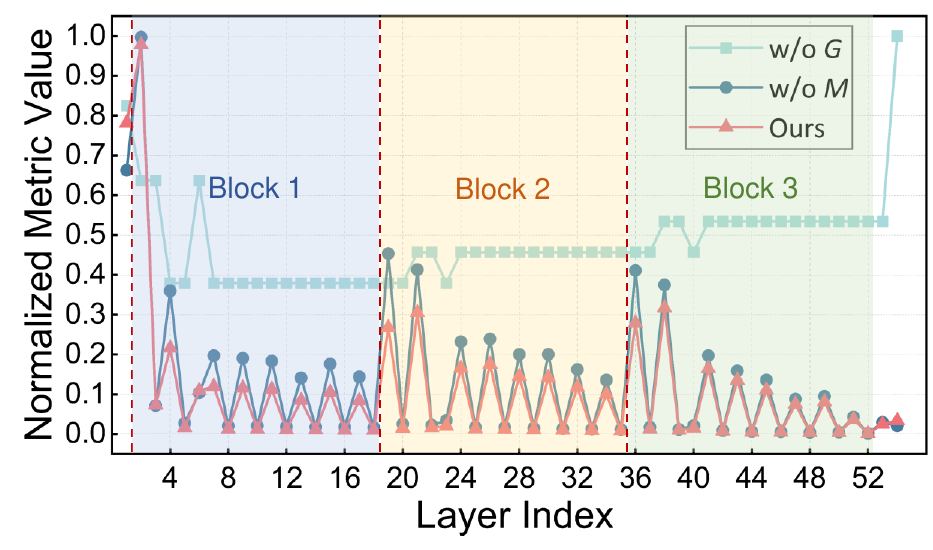}
    \caption{Visualization of normalized importance metrics. Experiments are implemented using WideResNet-28 on CIFAR10-C.
    }
    \label{fig_visualization}
\end{figure}

\vspace{-0.15in}
\paragraph{\namewithspace on ACDC}
The evaluation of semantic segmentation is depicted in Table \ref{tab_segmentation}.
Interestingly, all the chosen TTA baselines fail to surpass the accuracy of \textbf{Source} on DeeplabV3+.
This could potentially be attributed to the intricate nature of pixel-level prediction tasks and inaccurate pseudo-labels for adaptation, which may result in substantial error accumulation during TTA~\cite{lee2023towards}.
For Segformer-B5, \textbf{CoTTA} achieves the highest accuracy but incurs a substantial activation memory cost.
\textbf{MECTA} fails to improve accuracy during adaptation because it is designed based on BN layers, of which Segformer has only one, highlighting its limitations when applied to transformer-based architectures.
Notably, the cache size reported for \textbf{MECTA} and \name\ is the same as \textbf{Source} due to their forward propagation's activation memory usage exceeding that of backpropagation.

\vspace{-0.1in}
\paragraph{Summary} The above evaluation demonstrates that our method, \name, effectively balances memory cost and accuracy for FTTA without relying on specific architectures or modifications to the original training procedure. \namewithspace achieves a significant improvement in accuracy, as the layer-specific dynamic activation sparsity encourages adaptation in layers with higher accuracy contribution while suppressing unnecessary updates in less critical layers, thereby reducing error accumulation~\cite{lee2023surgical}.

\subsection{Empirical Study}
\label{subsec_ablation}
\paragraph{Effectiveness of Dynamic Activation Sparsity}
To demonstrate the effectiveness of dynamic activation sparsity, we compare \name\ with the method utilizing global static sparsity~\cite{jiang2022back,chen2023dropit}.
Figure \ref{fig_samecache} presents TTA accuracy under different global pruning ratios and reports pruning ratios of \namewithspace using the value of the global pruning ratio that produces equal activation memory cost.
The results demonstrate that \name\ achieves higher accuracy under the same cache size compared to static activation sparsity, as layer-wise dynamic sparsity captures the varying learning abilities of different layers to acquire knowledge and their different contribution to accuracy gains in a data-sensitive manner.

\begin{table}
\caption{Performance on Jetson Xavier NX. The experiments are conducted using ResNeXt-29 on CIFAR-100-C. GFLOPs and Latency refer to the number of $10^9$ floating-point operations and the time required to process a batch during FTTA.}
\Huge
\resizebox{\linewidth}{!}{ 
\begin{tabular}{l|c|c|c|c}
\hline
Methods & Mean (\%) $\downarrow$ & Cache (MB) $\downarrow$ & GFLOPs $\downarrow$ & Latency (ms) $\downarrow$ \\ \hline
BN-stat~\cite{schneider2020improving} & 44.0 & 8.0 & 8.4 & 25.4 \\ \hline
TENT~\cite{wang2021tent}    & 41.9 & 109.0 & \textbf{16.3} & \textbf{65.6} \\
CoTTA~\cite{wang2022continual}   & \textbf{37.5} & 216.0 & 259.2 & 522.6 \\
EcoTTA~\cite{song2023ecotta}  & 39.6 & 42.0 & 19.1 & 80.4 \\
MECTA~\cite{hong2023mecta}   & 38.2 & 22.8 & \underline{16.4} & \underline{66.5} \\ \hline
GC~\cite{chen2016training}      & 45.2 & 62.1 & 29.3 & 118.2 \\ \hline
\name & \underline{37.8} & \underline{22.4} & 28.3 & 117.4 \\
\namewithspace (BN) & 37.9 & \textbf{19.9} & 18.5 & 78.8 \\ \hline
\end{tabular}
}
\label{tab_computation}%
\end{table}

\vspace{-0.1in}
\paragraph{Effectiveness of Metric Components}
To evaluate the impact of $G$ (Eq.\eqref{equ_gradientimportance}) and $M$ (Eq.\eqref{equ_memoryefficiency}) as components of $I$ on the overall performance of \name, we record the accuracy and cache size by gradually removing each of the two metrics, as shown in Table \ref{tab_ablation}.
From the results, it is evident that the combination of $G$ and $M$ for determining layer-specific sparsity ratios achieves a state-of-the-art balance between memory cost and accuracy during FTTA. 

To more intuitively observe the impact of metric components, we visualize the normalized importance metrics using WideResNet-28 on CIFAR10-C, as shown in Figure \ref{fig_visualization}.
We can see that the first few layers within each block exhibit higher gradient importance values, indicating their greater contribution to accuracy.
Additionally, The activation memory importance of deeper layers is slightly higher than that of shallow layers, as their activation size is typically smaller (see Eq.~\eqref{equ_memoryefficiency}) due to downsampling.
Finally, the metric $I$ (i.e., ``Ours'') takes into account both the accuracy contribution and activation memory efficiency for informed decisions about pruning activations.

\vspace{-0.1in}
\paragraph{Analysis of Real-world Evaluation}
We also evaluate the methods by deploying them on Jetson XAVIER NX\footnote{\url{https://www.nvidia.cn/autonomous-machines/embedded-systems/jetson-xavier-nx/}}. The results are obtained using ResNeXt-29 on CIFAR-100-C with a test batch size of 8, as shown in Table \ref{tab_computation}.
Besides, in both the baselines (except MECTA) and our method, we calibrate the source and target statistics for BN layers to mitigate unreliable statistic estimation under small batch sizes~\cite{schneider2020improving}.
\namewithspace randomly selects a sample for layer importance calculation ($\S$ \ref{subsec_workflow}).
From the results, we observe that, our method achieves comparable accuracy and optimal cache reduction compared to the baselines, while maintaining acceptable computational costs and latency.
Meanwhile, we incorporate gradient checkpointing (GC)~\cite{chen2016training} for comparison.
Following the setup~\cite{hong2023mecta}, GC treats each residual block as the minimal unit.
Notably, GC does not alter activations for gradient calculation but temporarily discards them.
In constrast, our dynamic activation sparsity reduces activation memory and improves accuracy by precisely regulating both activation memory consumption and weight gradients across layers during FTTA.

\section{Conclusion}
\label{sec_conclusion}

This paper presents \name, a method aimed at reducing memory cost in FTTA while maintaining comparable accuracy, independent of specific architectures or alterations to the original training process. Considering inter-layer differences, \namewithspace leverages a novel dynamic activation sparsity strategy that prunes activations at layer-specific and adaptive ratios in a data-sensitive manner, precisely regulating both activation memory consumption and weight gradients across different layers during adaptation. Experimental results show that \namewithspace achieves SOTA balance between accuracy and memory across various datasets, architectures and tasks.
This research is expected to promote the robust deployment of deep models on resource-constrained devices in changing environments.

\vspace{-0.1in}
\paragraph{Acknowledgements.}

This work was supported by the National Science Fund for Distinguished Young Scholars (No.62025205), the National Natural Science Foundation of China (No.62472354, No.62302259, No.62432008), and Guangzhou-HKUST(GZ) Joint Funding Program (Grant No.2023A03J0008), Education Bureau of Guangzhou Municipality.

{
    \small
    \bibliographystyle{ieeenat_fullname}
    \bibliography{main}
}


\end{document}